\pdfoutput=1

\documentclass[11pt]{article}

\usepackage[]{acl}

\usepackage{times}
\usepackage{latexsym}
\usepackage{marginnote} 
\usepackage{xcolor} 
\usepackage{color,soul}
\setul{0.5ex}{0.3ex}
\setulcolor{blue}
\usepackage[T1]{fontenc}

\usepackage[utf8]{inputenc}

\usepackage{microtype}

\DeclareRobustCommand{\hlorange}[1]{{\sethlcolor{orange}\hl{#1}}}
\DeclareRobustCommand{\hlpink}[1]{{\sethlcolor{pink}\hl{#1}}}
\DeclareRobustCommand{\hlyellow}[1]{{\sethlcolor{yellow}\hl{#1}}}
\DeclareRobustCommand{\hlgreen}[1]{{\sethlcolor{green}\hl{#1}}}

\title{On the Ethical Considerations of Text Simplification}

\author{Sian Gooding \\
 Department of Computer Science and Technology \\
  University of Cambridge \\
  \texttt{shg36@cam.ac.uk}}

\begin{document}
\maketitle
\begin{abstract}
This paper outlines the ethical implications of text simplification within the framework of assistive systems. We argue that a distinction should be made between the technologies that perform text simplification and the realisation of these in assistive technologies. When using the latter as a motivation for research, it is important that the subsequent ethical implications be carefully considered. We provide guidelines for the framing of text simplification independently of assistive systems, as well as suggesting directions for future research and discussion based on the concerns raised.
\end{abstract}

\section{Introduction}
Assistive technology refers to the devices used to support or aid those living with disabilities \cite{preston2003assistive}. The intent behind such technologies is to increase independence and maximise societal participation for individuals \cite{borg2011right}. 

There are many examples of assistive technology that rely on speech and natural language processing. For instance, sign language translation \cite{camgoz2018neural}, pronunciation adaptation for disordered speech \cite{sriranjani2015pronunciation} and synthesised voices for individuals with vocal disabilities \cite{veaux2013towards}. \textit{Text simplification} is an area of natural language processing concerned with the simplification of textual information and is often recognised as having assistive applications. Prior research in text simplification posits that such technology may be beneficial for audiences with reading difficulties or a range of disabilities such as dyslexia, aphasia or deafness. 

However, currently the algorithms designed for text simplification are considered in isolation from their assistive applications, and there is subsequently little discussion on the ethical implications for the intended users. Text simplification research is often motivated by highlighting the audiences that could benefit from such tools, thereby coupling the technology with the assistive applications. Framing text simplification via the implications for assistive technology means that the ethical considerations cannot be easily separated from the technology used to generate the result. An issue which is commonly acknowledged in the assistive technology literature \cite{niemeijer2010ethical}.

There are many potential benefits of text simplification embedded in assistive technology, and both for service providers and service users, there are also a number of ethical
issues that must be considered. In this paper, we will discuss the ethical considerations that arise from the embedding of text simplification within assistive technologies. Our aim is to encourage the discussion and consideration of these issues, as well as inform the design decisions of future assistive technologies that incorporate text simplification. 

\section{Background}
\begin{table}[t]
\centering
\begin{tabular}{l}
\hline
\multicolumn{1}{|l|}{\begin{tabular}[c]{@{}l@{}}\small{Some philatelists say the committee that helps the} \\ \small{postmaster general pick new stamps is favoring pop} \\ \small{celebrities and fictional characters over cultural sites} \\ \small{and historical figures, undermining a long tradition.}\end{tabular}}                                                                                     \\ \hline
\multicolumn{1}{c}{$\downarrow$}                                                                                                        \\ \hline
\multicolumn{1}{|l|}{\begin{tabular}[c]{@{}l@{}}\small{Some philatelists (\hlpink{as stamp collectors are known}) say} \\ \small{the committee that helps pick new stamps is favoring} \\ \small{pop \hlorange{stars} and fictional characters. Such choices mean}\\ \small{that cultural sites and historical figures are appearing} \\ \small{less often\hlyellow{.} \hlgreen{They say} this results in the undermining of} \\ \small{a long tradition.}\end{tabular}} \\ \hline
\end{tabular}
\caption{Example of manually simplified sentence from the Newsela Dataset \cite{xu:2015:TACL}}
\label{tab:simpl_example}
\end{table}

Complete textual simplification requires many types of transformations which can be grouped into three categories: syntactic, lexical and conceptual \cite{siddharthan2014survey}. Table \ref{tab:simpl_example} illustrates a range of simplification operations from these different categories, a description of these is as follows:

\setulcolor{orange}\ul{Lexical simplification} is concerned with reducing the complexity of words within a text \cite{paetzold2017survey, gooding-kochmar-2019-recursive}. In lexical simplification, complex words are identified and replaced with simpler alternatives. We observe an example of lexical simplification with the case of \textit{celebrities} being simplified to \textit{stars}. 

\setulcolor{yellow}\ul{Syntactic simplification} aims to reduce the grammatical complexity of text by simplifying the syntactical structures. Examples of such transformations include the conversion of text from passive to active voice and dis-embedding relative clauses \cite{siddharthan2006syntactic}. In our example, multiple syntactic simplifications have taken place. One such simplification occurs where the subordinated clause `\textit{...undermining a long tradition}' has been split into a separate sentence. Syntactic simplification often requires \setulcolor{green}\ul{discourse preserving edits} to maintain the coherence and cohesion of simplified text. For instance, the addition of `\textit{They say...}' is necessary to convert the original relative clause into a grammatically correct and coherent sentence.

Finally, \setulcolor{pink}\ul{conceptual simplification} focuses on the simplification of ideas or concepts within text. The example shows how the concept of \textit{philatelist} has been simplified by providing an explanation of the term. This simplification technique is commonly referred to as elaboration, as the meaning of the concept has been elaborated on \cite{siddharthan2006syntactic}. Often, this strategy is used in cases where no alternative synonym would suffice, for instance with named entities. 

Both syntactic and conceptual simplification contain parallels with the research area of \textit{text summarization} as omitting peripheral or inappropriate information, as well as distilling complex concepts, is relevant for both. However, in simplification these processes can increase the length of the original text, whereas in summarization the goal is to constrain the length of the resulting summary. 

In automatic text simplification, the aim is to transform text using the aforementioned operations, to allow individuals with differing comprehension levels access. This requires a fundamental understanding of what factors contribute to text complexity for differing audiences \cite{gooding-etal-2021-word}. 

Early approaches to automated simplification were largely rule-based systems~\cite{canning-EtAl:2000,carroll-EtAl:1998,siddharthan:2006}, with many prioritising syntactic operations, such as sentence splitting, deletion or reordering. However, some work combined lexical simplification with syntactic operations~\cite{coster-kauchak:2011:W11-1601,kauchak:2013:P13-1151,zhu-EtAl:2010:C10-1152}. In recent years simplification has been viewed as a monolingual translation task~\cite{kauchak:2013:P13-1151,zhang:2017:D17-1062,zhu-EtAl:2010:C10-1152}. These systems perform a number of simplification operations at once by aiming to translate \textit{complex English} to \textit{simple English}. Initial approaches attempt this with phrase-based machine translation \cite{coster-kauchak:2011:W11-1601,wubben2012sentence} while subsequent work has focused on neural machine translation techniques \cite{nisioi2017exploring,zhang:2017:D17-1062, shardlow2019neural, dong-etal-2019-editnts}.

\section{Risks and Harms}
In this section we outline and discuss the potential risks and harms that arise from the integration of text simplification within assistive technology. 
\subsection{Intended Audience}

\begin{table*}[t]
\begin{tabular}{p{8cm}|p{2cm}|p{2cm}|p{2cm}}
\textit{Audience outline}                                                                                                                                                                                                               & \textit{Datasets}                                                      & \textit{Evaluation}                                              & \textit{Venue}                   \\ \hline
\textit{\small{(1) ...such as children, people with low education, people who have reading disorders or dyslexia, and non-native speakers of the language.}}                                   & \begin{tabular}[c]{@{}l@{}}\small{\textsc{Newsela}}\\ \small{\textsc{WikiLarge}}\\ \small{\textsc{Biendata}}\end{tabular} & \small{Automatic}                                                        & \small{\emph{ACL 2021}} \\ \hline
\textit{\small{(2) It provides reading assistance to children \cite{kajiwara2013selecting}, non-native speakers \cite{petersen2007text,pellow2014open, paetzold2016lexical} and people with reading disabilities \cite{rello2013impact}}}                                                                    & \small{\textsc{Newsela}}                                                                & \begin{tabular}[c]{@{}l@{}}\small{Automatic +}\\ \small{5 workers}\end{tabular}  & \small{\emph{NAACL 2021}} \\ \hline
\textit{\small{(3) It can provide convenience for non-native speakers \cite{petersen2007text,glavavs2015simplifying,paetzold2016unsupervised,rello2013impact}, non-expert readers \cite{elhadad2007mining, siddharthan2010reformulating} and children \cite{de2010text,kajiwara2013selecting}}}                                                                                        & \begin{tabular}[c]{@{}l@{}}\small{\textsc{D-Wikipedia}}\\ \small{\textsc{Newsela}}\end{tabular}          & \begin{tabular}[c]{@{}l@{}}\small{Automatic +} \\ \small{3 workers}\end{tabular} & \small{\emph{EMNLP 2021}}         \\ \hline
\textit{\small{(4) ...to children \cite{de2010text,kajiwara2013selecting}, people with language disabilities like aphasia \cite{carroll1998practical,carroll1999simplifying,devlin2006helping}, dyslexia \cite{rello2013simplify, rello2013impact}, or autism \cite{evans2014evaluation}; non-native \cite{petersen2007text, paetzold2015reliable,paetzold2016understanding,pellow2014open} English speakers, and people with low literacy skills or reading ages.}} & \begin{tabular}[c]{@{}l@{}}\small{\textsc{WikiSmall}} \\ \small{\textsc{WikiLarge}}\end{tabular}         & \small{Automatic}                                                        & \small{\emph{BEA 2021}}    
\end{tabular}
\caption{Examples of paper introductions outlining audiences benefiting from text simplification, alongside the evaluation techniques and venue, specific paper references are included in Appendix \ref{sec:appendix}.}
\label{tab:preamble}
\end{table*}

As with many areas of research, the field of text simplification has converged on a partially boilerplate preamble outlining a set of motivations. Table \ref{tab:preamble} features extracts taken from a sample of recent text simplification papers. These papers were sampled by searching the ACL anthology for the term \textit{text simplification} and ordering by most recent. We look specifically at sections outlining the audiences said to benefit from text simplification as a whole. Below, we consider the ethical implications of citing such audiences as a motivation for text simplification.

\subsubsection{The Homogeneity Effect}
As shown in Table \ref{tab:preamble}, the audiences stated to benefit from text simplification are often listed together, namely non-native speakers, children, people with low literacy skills, people with reading disabilities or disabilities generally. Based on this, a reader may be given the impression that general purpose text simplification works adequately for all of the stated groups. Whereas in fact, there is evidence to show that text simplification may not be effective for second language learners \cite{young1999linguistic}, that alternative strategies to simplification can be most effective for dyslexia \cite{rello2013simplify} and that automated text simplification cannot simplify content to a low enough level for children \cite{de2010text}.

Framing the benefit of text simplification as net positive for all groups can have consequences for the development of assistive technology, as merging the audiences serves to diminish the sensitive differences in needs for these groups. Even for specific audiences, such as children, there is a consensus that the homogeneous grouping of reading ability can have detrimental outcomes for learning \cite{schumm2000grouping}.

A further complication, is that the references commonly used in support of text simplification (for specific audiences) are often more nuanced than stated. For instance, the work of \citet{rello2013simplify} is commonly cited as showing the benefit of text simplification for dyslexic readers. However, this paper demonstrates that the most effective strategy to help dyslexic readers with difficult words, is to provide a range of synonyms for the word, and not to simplify the original. Furthermore, the work of \citet{carroll-EtAl:1998} and \citet{carrollEtAl:1999:EACL} is put forward as evidence for the utility of simplification for individuals with aphasia. However, both of these works outline the proposal for a simplification system targeted for aphasia and propose to evaluate the effectiveness of such a system in future work. A final example, used in support of text simplification for children is a paper by \citet{de2010text}. However, the paper finds that even using lexical and syntatic simplification, it was not possible to reduce the reading difficulty enough for children.

\subsubsection{Datasets and Evaluation}
\label{sec:audience}
Text simplification has many subtleties, as what would be a valid simplification for one reader may not be appropriate for another \cite{xu:2015:TACL}. For instance, it has been shown that the factors contributing to word complexity vary depending on the first language and proficiency level of a reader \cite{gooding-etal-2021-word}. The subjective nature of text simplification means that system evaluation is difficult. Furthermore, as there is not one `ground truth' for simplification, the efficacy of automatic evaluation measures is limited. Prior work on the development and evaluation of simplification systems has given little consideration to the target reader population \cite{xu:2015:TACL}. 

As exemplified in Table \ref{tab:preamble}, current work on text simplification typically relies on automatic evaluation, with the occasional use of human evaluation. When considering the approach to human evaluation, most work does not specify what “being simpler” entails, and trusts human judges to use their own understanding of the concept \cite{alva2021suitability}. It is also currently not standard practice to include the demographic information of the workers. When using human judgements as a measure of simplification quality, it is important to include relevant information on the demographic background, so that valid conclusions can be drawn about which target population may benefit from the system. Additionally, the concept of what constitutes adequate simplification needs to be precise if the system is aimed for a specialised audience. 

The datasets commonly used to train text simplification systems (i.e. Newsela and Simple Wikipedia) have drawbacks such as poor alignment, lack of simplicity and not being tailored for a specific audiences \cite{xu:2015:TACL}. In text simplification, it is important to discuss the limitations of the data so that the suitability of such systems for specialised groups is clearly recognised. 

In summary, when claiming the benefits of text simplification for specific audiences, it is crucial that the needs of these groups are understood. This is especially the case when emphasising the benefit of such technology for disabled groups. The development of assistive technology is downstream from research, and therefore being clear about the suitability and limitations of the technology for differing audiences helps to avoid poorly suited assistive technology solutions being developed. 

\subsection{Meaning Distortion}
There are multiple genres of text where access is highly important, such as healthcare information or political materials. The benefits of simplifying such content have been shown, for example simplifying text in health care improves understanding of information regardless of health literacy level \cite{kim2015simplification}. Furthermore, the complexity of language matters for voters’ perceptions of political parties and their positions \cite{bischof2018simple}. 

The benefit of text simplification in such cases is apparent, as is the need to ensure the meaning of such text is preserved and that no errors are introduced. A drawback to current automated text simplification systems is that the subtleties of meaning intended by the author may be diluted, if not lost altogether \cite{chandrasekar1996motivations}. For example, \citet{shardlow2019neural} found that fully automated approaches omitted 30\% of critical information when used to simplify clinical texts.  For these types of domains, instead of fully-automated approaches, interactive text simplification tools are better suited to generate more efficient and higher quality simplifications \cite{kloehn2018improving}.

The link between factual correctness and natural language generation has been considered for multiple domains such as summarization \cite{cao2018faithful}, data to document generation \cite{wiseman-etal-2017-challenges} and dialog generation \cite{ shuster-etal-2021-retrieval-augmentation}. However, this is a relatively underexplored area for text simplification and is currently not incorporated into the evaluation of such systems. 

Encouraging further discussion on this limitation of text simplification is necessary. Especially when we consider the downstream applications of assistive technology for critical consumer information. 

\subsection{Paternalism}
\label{sec:pat}
There are choices made in the design characteristics of assistive technology that can affect the degree of independence, privacy and participation that are possible \cite{lenker2013consumer}. These decisions have real world impact for the users and thus warrant careful consideration.

The process of text simplification involves an understanding of \textit{what} is difficult and \textit{how} best to simplify it. There are two approaches when deciding \textit{what} should be simplified. The first, involves including the reader in the loop -- either implicitly or explicitly. Relying on user signal to identify areas for simplification has its own set of ethical concerns which are discussed in $\S$ \ref{sec:priv}. The second approach, is performing general level simplification with end-to-end systems. In fact, the majority of current work in text simplification is now data-driven and performs simplification in a `black-box' fashion \cite{sikka2020survey}. One of the concerns for such systems, is that they learn operations based on the simplification choices made in the data they are trained with. As outlined in Section \ref{sec:audience}, most of the data used to train text simplification systems is not audience specific \cite{xu:2015:TACL}.

Integrating general purpose simplification systems into assistive technologies has a range of potential problems. For instance, it raises the issue of ``paternalism" which is the interference of a state or individual in relation to another person \cite{martin2007ethical}. The relationship between paternalism and assistive technology is widely acknowledged, as design decisions made on behalf of a user can be problematic if they override the autonomy of the individual
\cite{martin2007ethical, martin2010assistive}. In the case of text simplification, not allowing the individual the choice of what they would want simplified restricts their autonomy.

Furthermore, assistive technologies should contribute to growth and independence for individuals. The goals of text simplification are to make textual information accessible to a range of different audiences. However, the question of whether such systems should support learning is rarely discussed. One concern with text simplification within assistive systems, is that it would prevent the exposure to new terms and concepts thereby encouraging learning stagnation. 

In summary, the design decisions pertaining to what content is simplified have ethical implications for the user. Removing the individual from the decision process can reduce the person's autonomy, and not allowing exposure to new and unfamiliar terms limits learning opportunities, subsequently reducing the user's independence. 

\subsection{Privacy and Security}
\label{sec:priv}
As outlined in Section \ref{sec:pat}, an effective approach for text simplification in assistive technology is to include the user in the decision of what is simplified. Prior research has shown that eye-tracking \cite{berzak-etal-2018-assessing} and scroll-based interactions \cite{gooding-etal-2021-predicting} correlate with text understanding. As such, these implicit techniques can be used to gain an insight into what the reader is finding difficult. The reader can also be explicitly asked to select text that they would like to be simplified, for instance by selecting words that are difficult for them \cite{devlin2006helping, paetzold2016anita}. 

Adaptive text simplification is advantageous and provides autonomy and learning opportunities for the user. However, the information about the areas a user finds difficult is highly sensitive, and there is a responsibility to ensure that such information is stored securely. 

To protect the privacy of the user, the aim of the assistive technology and the way it is used by service providers or care organisations must be clear. Moreover, how personal data will be handled must be described explicitly in a privacy statement and communicated to the user \cite{martin2010assistive}. 

This is a clear example of how viewing text simplification through the paradigm of assistive technology yields more nuanced ethical considerations. We believe it would be beneficial to encourage the discourse on such aspects in the text simplification literature. 

\section{Going Forward}
We suggest that papers focusing on general purpose text simplification should de-couple the motivations from specific audiences with disabilities. An example of a general purpose motivation by \citet{nisioi2017exploring} is as follows:
\begin{center}
\textit{Automated text simplification (ATS) systems are meant to transform original texts into different (simpler) variants which would be understood by wider audiences and more successfully processed by various NLP tools.}
\end{center}
Alternatively, if discussing the different groups of users who may benefit from text simplification, being clear about the specific strategies that work for these audiences is critical. Additionally, it is worth acknowledging that when framing a system using a target demographic, it is appropriate that the system is tested with that target group. For human evaluation generally, it is highly beneficial to report the demographic statistics, as this allows an insight into which types of audiences the system may work well for. 

Finally, it is important to be forthright about the current limitations of both the data and evaluation techniques used in automatic text simplification. Whilst great improvements are being made in this area, these systems are still far from perfect and this needs to be taken into account when judging the suitability of systems for assistive technology. 
\section{Conclusion}
Assistive technologies can dramatically affect the lives of those who rely on them, and it is important to understand the potential ethical concerns -- especially as such technologies can impact vulnerable populations. In this paper, we discuss a set of potential issues that arise from the embedding of text simplification within assistive technologies. 

Our aim in this work is to encourage further discussion on how the design decisions of text simplification algorithms can have the potential to impact future users of assistive technology.
\bibliography{anthology,custom}
\bibliographystyle{acl_natbib}

\appendix

\section{Appendix}
\label{sec:appendix}
Paper references from Table \ref{tab:preamble}:
\begin{enumerate}
    \item Explainable Prediction of Text Complexity: The Missing Preliminaries for Text Simplification by \citet{garbacea-etal-2021-explainable}
    \item by Controllable Text Simplification with Explicit Paraphrasing by \citet{maddela-etal-2021-controllable}
    \item Document-Level Text Simplification: Dataset, Criteria and Baseline by \citet{sun-etal-2021-document}
    \item Text Simplification by Tagging by \citet{omelianchuk-etal-2021-text}
\end{enumerate}

\end{document}